\documentclass[12pt,a4paper,oneside]{report}             
\usepackage{ifxetex}
\usepackage{ifluatex}
\newif\ifxetexorluatex 
\ifnum 0\ifxetex 1\fi\ifluatex 1\fi>0
   \xetexorluatextrue
\fi

\ifxetexorluatex
  \usepackage{fontspec}
\else
  \usepackage[T1]{fontenc}
  \usepackage[utf8]{inputenc}
  \usepackage[lighttt]{lmodern}
\fi

\usepackage[english,magyar]{babel} 

\usepackage{amsfonts,amsmath,amssymb} 
\usepackage{booktabs} 
\usepackage{graphicx}

\usepackage{anysize}
\usepackage{setspace} 

\usepackage[unicode]{hyperref} 
\usepackage{xcolor}
\usepackage{listings} 

\usepackage[amsmath,thmmarks]{ntheorem} 

\usepackage[hang]{caption}

\newcommand{\selecthungarian}{
	\selectlanguage{magyar}
	\setlength{\parindent}{2em}
	\setlength{\parskip}{0em}
	\frenchspacing
}

\newcommand{\selectenglish}{
	\selectlanguage{english}
	\setlength{\parindent}{0em}
	\setlength{\parskip}{0.5em}
	\nonfrenchspacing
	\renewcommand{\figureautorefname}{Figure}
	\renewcommand{\tableautorefname}{Table}
	\renewcommand{\partautorefname}{Part}
	\renewcommand{\chapterautorefname}{Chapter}
	\renewcommand{\sectionautorefname}{Section}
	\renewcommand{\subsectionautorefname}{Section}
	\renewcommand{\subsubsectionautorefname}{Section}
}

\usepackage[numbers]{natbib}
\usepackage{xspace}

\newcommand{\normalcode}[1]{\code{\normalsize #1}}

\newcommand{\figuresize}{0.65}

\newcommand{\vikszerzoVezeteknev}{Hevesi}
\newcommand{\vikszerzoKeresztnev}{Gergely}

\newcommand{\vikkonzulensAMegszolitas}{}
\newcommand{\vikkonzulensAVezeteknev}{Futóné Papp}
\newcommand{\vikkonzulensAKeresztnev}{Dorottya}

\newcommand{\vikkonzulensBMegszolitas}{}
\newcommand{\vikkonzulensBVezeteknev}{}
\newcommand{\vikkonzulensBKeresztnev}{}

\newcommand{\vikkonzulensCMegszolitas}{}
\newcommand{\vikkonzulensCVezeteknev}{}
\newcommand{\vikkonzulensCKeresztnev}{}
\newcommand{\vikcim}{Machine learning-based malware detection for IoT devices using control-flow data} 
\newcommand{\viktanszek}{\bmehit} 
\newcommand{\vikdoktipus}{\bsc} 
\newcommand{\vikmunkatipusat}{szakdolgozatot} 

\newcommand{\szerzoMeta}{\vikszerzoVezeteknev{} \vikszerzoKeresztnev} 

\newcommand{\bme}{Budapest University of Technology and Economics}
\newcommand{\vik}{Faculty of Electrical Engineering and Informatics}

\newcommand{\bmehit}{Department of Networked Systems and Services}

\newcommand{\keszitette}{Author}
\newcommand{\konzulens}{Advisor}

\newcommand{\bsc}{Bachelor's Thesis}

\newcommand{\pelda}{Example}
\newcommand{\definicio}{Definition}
\newcommand{\tetel}{Theorem}

\newcommand{\bevezetes}{Introduction}
\newcommand{\koszonetnyilvanitas}{Acknowledgement}

\newcommand{\szerzo}{\vikszerzoKeresztnev{} \vikszerzoVezeteknev}
\newcommand{\vikkonzulensA}{\vikkonzulensAMegszolitas\vikkonzulensAKeresztnev{} \vikkonzulensAVezeteknev}
\newcommand{\vikkonzulensB}{\vikkonzulensBMegszolitas\vikkonzulensBKeresztnev{} \vikkonzulensBVezeteknev}
\newcommand{\vikkonzulensC}{\vikkonzulensCMegszolitas\vikkonzulensCKeresztnev{} \vikkonzulensCVezeteknev}

\newcommand{\selectthesislanguage}{\selectenglish}

\marginsize{35mm}{25mm}{15mm}{15mm}

\hypersetup{
    unicode=true,              
    pdftitle={\vikcim},        
    pdfauthor={\szerzoMeta},    
    pdfsubject={\vikdoktipus}, 
    pdfcreator={\szerzoMeta},   
    pdfproducer={},    
    pdfkeywords={},    
    pdfnewwindow=true,         
    colorlinks=true,           
    linkcolor=black,           
    citecolor=black,           
    filecolor=black,           
    urlcolor=black             
}

\definecolor{lightgray}{rgb}{0.95,0.95,0.95}
\theoremstyle{plain}
\theoremseparator{.}

\theoremseparator{.}
\theorembodyfont{\upshape}
\theoremsymbol{{\large \ensuremath{\centerdot}}}

\theoremseparator{.}

\newcommand{\code}[1]{{\upshape\ttfamily\scriptsize\indent #1}}
\newcommand{\doi}[1]{DOI: \href{http://dx.doi.org/\detokenize{#1}}{\raggedright{\texttt{\detokenize{#1}}}}} 

\author{\vikszerzo}
\title{\viktitle}

\usepackage[bottom]{footmisc}
\usepackage{multirow}
\usepackage{float}

\begin{document}

\pagenumbering{gobble}

\selectthesislanguage

\hypersetup{pageanchor=false}
\begin{titlepage}
\begin{center}
\includegraphics[width=60mm,keepaspectratio]{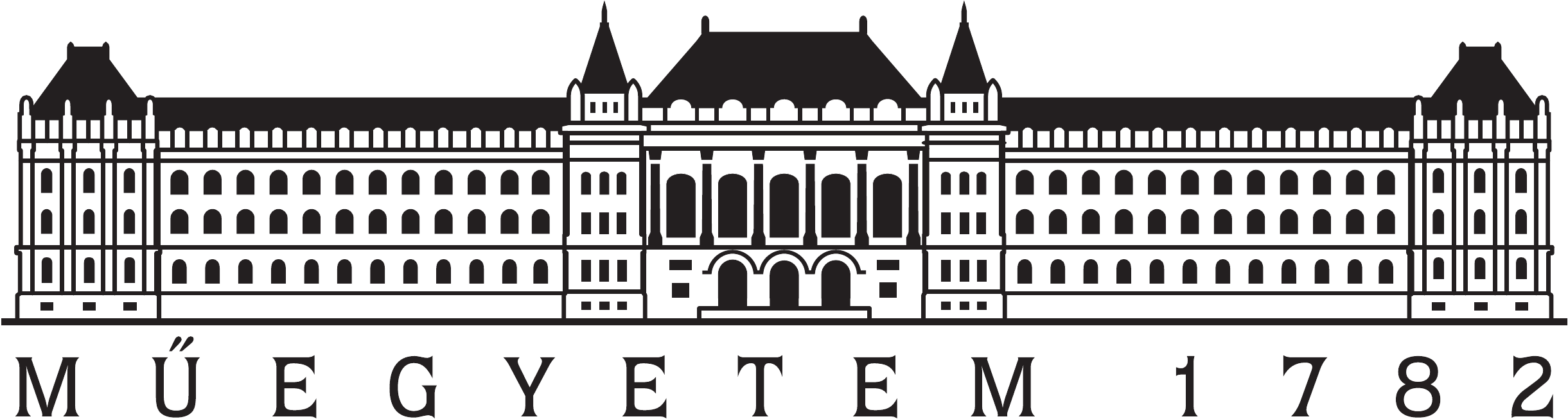}\\
\vspace{0.3cm}
\textbf{\bme}\\
\textmd{\vik}\\
\textmd{\viktanszek}\\[5cm]

\vspace{0.4cm}
{\huge \bfseries \vikcim}\\[0.8cm]
\vspace{0.5cm}
\textsc{\Large \vikdoktipus}\\[4cm]

{
	\renewcommand{\arraystretch}{0.85}
	\begin{tabular}{cc}
	 \makebox[7cm]{\emph{\keszitette}} & \makebox[7cm]{\emph{\konzulens}} \\ \noalign{\smallskip}
	 \makebox[7cm]{\szerzo} & \makebox[7cm]{\vikkonzulensA} \\
	  & \makebox[7cm]{\vikkonzulensB} \\
	  & \makebox[7cm]{\vikkonzulensC} \\
	\end{tabular}
}

\vfill
{\large \today}
\end{center}
\end{titlepage}
\hypersetup{pageanchor=false}


\tableofcontents\vfill

\selectlanguage{magyar}
\pagenumbering{gobble}
\begin{center}
\large
\textbf{HALLGATÓI NYILATKOZAT}\\
\end{center}

Alulírott \emph{\vikszerzoVezeteknev{} \vikszerzoKeresztnev}, szigorló hallgató kijelentem, hogy ezt a \vikmunkatipusat{} meg nem engedett segítség nélkül, saját magam készítettem, csak a megadott forrásokat (szakirodalom, eszközök stb.) használtam fel. Minden olyan részt, melyet szó szerint, vagy azonos értelemben, de átfogalmazva más forrásból átvettem, egyértelműen, a forrás megadásával megjelöltem.

Hozzájárulok, hogy a jelen munkám alapadatait (szerző(k), cím, angol és magyar nyelvű tartalmi kivonat, készítés éve, konzulens(ek) neve) a BME VIK nyilvánosan hozzáférhető elektronikus formában, a munka teljes szövegét pedig az egyetem belső hálózatán keresztül (vagy autentikált felhasználók számára) közzétegye. Kijelentem, hogy a benyújtott munka és annak elektronikus verziója megegyezik. Dékáni engedéllyel titkosított diplomatervek esetén a dolgozat szövege csak 3 év eltelte után válik hozzáférhetővé.

\begin{flushleft}
\vspace*{1cm}
Budapest, \today
\end{flushleft}

\begin{flushright}
 \vspace*{1cm}
 \makebox[7cm]{\rule{6cm}{.4pt}}\\
 \makebox[7cm]{\emph{\vikszerzoVezeteknev{} \vikszerzoKeresztnev}}\\
 \makebox[7cm]{hallgató}
\end{flushright}
\thispagestyle{empty}

\vfill
\clearpage
\thispagestyle{empty} 

\selectthesislanguage
\pagenumbering{roman}
\setcounter{page}{1}

\selecthungarian

\chapter*{Kivonat}\addcontentsline{toc}{chapter}{Kivonat}
A beágyazott eszközök olyan speciális eszközök, amelyeket egy vagy csak néhány célra terveztek. Gyakran egy nagyobb rendszer részei, vezetékes vagy vezeték nélküli kapcsolaton keresztül. Azokat a beágyazott eszközöket, amelyek az interneten keresztül más számítógépekhez vagy beágyazott rendszerekhez kapcsolódnak, a dolgok internetének (röviden IoT) nevezzük.

Széleskörű használatuk és elégtelen védelmük miatt ezek az eszközök egyre inkább rosszindulatú támadások célpontjává válnak. A vállalatok gyakran a gyártási költségekkel spórolnak vagy félrekonfigurálnak, amikor ezeket az eszközöket gyártják. Ez lehet a szoftverfrissítések hiánya, a nyitva hagyott portok vagy a tervezésből adódó biztonsági hibák. Bár ezek az eszközök nem rendelkeznek akkora teljesítménnyel, mint egy hagyományos számítógép, nagy számuk miatt alkalmasak botnetek kialakítására. Más típusú IoT-eszközök akár egészségügyi problémákat is okozhatnak, hiszen még pacemakerek is csatlakoznak az internetre. Ez azt jelenti, hogy megfelelő védelem nélkül akár az emberek elleni irányított támadások is lehetségesek.

A szakdolgozatom célja, hogy gépi tanulási algoritmusok és visszafejtő programok segítségével jobb biztonságot nyújtson ezeknek az eszközöknek. A szakdolgozatban egy szétszerelő részből és egy neurális hálózatból álló rendszert mutatok be, amit egy ARM malware és router firmware képekből álló adatbázison teszteltem. A szétszerelő ebben az esetben egy hívásgráf reprezentációt hoz létre, amely egy neurális hálózat bemenete. Ez a gráf a betanítási folyamat során címkézésre kerül aszerint, hogy rosszindulatú vagy jóindulatú mintáról van-e szó. 

\vfill
\selectenglish

\chapter*{Abstract}\addcontentsline{toc}{chapter}{Abstract}
Embedded devices are specialised devices designed for one or only a few purposes. They are often part of a larger system, through wired or wireless connection. Those embedded devices that are connected to other computers or embedded systems through the Internet are called Internet of Things (IoT for short) devices.

With their widespread usage and their insufficient protection, these devices are increasingly becoming the target of malware attacks. Companies often cut corners to save manufacturing costs or misconfigure when producing these devices. This can be lack of software updates, ports left open or security defects by design. Although these devices may not be as powerful as a regular computer, their large number makes them suitable candidates for botnets. Other types of IoT devices can even cause health problems since there are even pacemakers connected to the Internet. This means, that without sufficient defence, even directed assaults are possible against people.

The goal of this thesis project is to provide better security for these devices with the help of machine learning algorithms and reverse engineering tools. Specifically, I study the applicability of control-flow related data of executables for malware detection. I present a malware detection method with two phases. The first phase extracts control-flow related data using static binary analysis. The second phase classifies binary executables as either malicious or benign using a neural network model. I train the model using a dataset of malicious and benign ARM applications.

\vfill
\selectthesislanguage
\newcounter{romanPage}
\setcounter{romanPage}{\value{page}}
\stepcounter{romanPage}

\pagenumbering{arabic}

\chapter{\bevezetes}

Embedded devices are small microcomputers that usually possess low computing capacity and power consumption. With the widespread of smart home appliances and Industry 4.0, they are often connected to the Internet for remote access and data collection. These embedded devices are called IoT devices. These are made for specific tasks, rather than universal computing like traditional computers. Nowadays this is one of the largest trends, that connects the virtual and physical world. Although some of them only control automated air conditioning or lights in our home, they often perform critical tasks in the enterprise world because of their reliability and hardware-implemented functions. The combination of these means, that they are essential parts of some industries.

IoT devices are typically not designed with security in mind, despite their widespread usage. The reason of insufficient defence can be anything from cost saving measures, lack of processing power or simply absence of attention. On vulnerability database sites, such as the Common Vulnerabilities and Exposures\footnote{\url{https://cve.mitre.org/} last accessed: 08/12/2021} (CVE) there are hundreds of known weaknesses in embedded devices. There is remote code execution in router firmwares using crafted IP payload in CVE-2021-41653\footnote{\url{https://cve.mitre.org/cgi-bin/cvename.cgi?name=CVE-2021-41653} last accessed: 08/12/2021}, inadequate encryption strength in CVE-2021-38464\footnote{\url{https://cve.mitre.org/cgi-bin/cvename.cgi?name=CVE-2021-38464} last accessed: 08/12/2021} or ARP poisoning in CVE-2021-29280\footnote{\url{https://cve.mitre.org/cgi-bin/cvename.cgi?name=CVE-2021-29280} last accessed: 08/12/2021}. The lack of security and connection to inner networks means they are often playing the role of gateways in security breaches. Their large numbers and close integration with our lives means they can often be used for botnet attacks or data collection on a massive scale. Considering these, it is becoming increasingly important to implement some kinds of defence against these attacks.

In the past few years, not only the number of attacks but their variety has also increased.  Attackers have also moved towards automating attacks using malware. Entire malware families, e.g., Tsunami/Kaiten and Mirai \cite{Tangled_IoT_malware}, developed from one type of malware and this made their detection much more difficult. These families are similar to the biological evolution, as they are also constantly evolving in reaction to the environment. As a result, the generally accepted method of hash checking has become obsolete, as even small changes in a malware sample's binary representation cause an entirely new hash. Traditionally, extensive human resources were allocated for deep malware analysis, however, the growing number of IoT malware increasingly makes this approach challenging and unscalable. This implies that we need methods for IoT malware detection. This has its own challenges such as dealing with low performance which makes it impossible to apply traditional methods. Attackers have developed a number of countermeasures to make detection challenging, such as code polymorphism, code encryption and packed code. Their increase in their number caused that we have a vast sample size for analysing their structure and behaviour.

Considering the also enormous improvement in computational capacity in recent years, all kinds of machine learning algorithms became available for malware detection \cite{The_rise_of_machine_learning_for_detection_and_classification_of_malware,Survey_of_machine_learning_techniques_for_malware_analysis,A_Survey_on_Malware_Detection_Using_Data_Mining_Techniques}. These work best when enough data is available, which means that the downside of this increase in malware can be the strength of machine learning. There are multiple types to choose from, depending on the nature of the problem, the computational power of the model building device and the client, and if we want feedback from the algorithm about the reasoning regarding its decision.
 
In this thesis project I created a malware detection framework that uses machine learning techniques for identifying harmful programs. The framework generates the call graphs of the input programs and after tagging its nodes uses them as features. These features then become the input vector for a neural network, for which I used a PyTorch framework\footnote{\url{https://pytorch.org} last accessed: 09/12/2021} implementation of the structure2vec algorithm~\cite{Struct2vec}.

Considering its widespread use in IoT devices, I used ARM malware and benign samples. Their combined number is 1.396, which contained 1.054 malware and 547 benign sample therefore I had to balance the dataset. Benign samples were extracted from D-Link and Ubiquiti firmware images I collected from the manufacturers' websites\footnote{\url{https://support.dlink.com} last accessed: 27/11/2021 and \url{https://www.ui.com/download} last accessed: 27/11/2021}. After the collection of benign samples and their libraries, with the help of the angr\footnote{\url{https://angr.io} last accessed: 24/11/2021} framework, I extracted the control-flow data in the format required by the implemented structure2vec algorithm. Although the algorithm was created to classify molecules, the concept was still the same: decide which class does a specific graph input belongs to. In the end, I evaluated and summarized the result of classification and propounded some ideas for future work projects. The structure of this thesis is the following:

\begin{itemize}
	\item Chapter \ref{ch:Literature} details the background knowledge required to understand the concept of the thesis.
	\item Chapter \ref{ch:Overview} gives an outline about the framework of the thesis project, which is detailed in Chapters \ref{ch:CF_recovery}, \ref{ch:Preprocessing}, and \ref{ch:Struct2vec}.
	\item Chapter \ref{ch:CF_recovery} explains the steps required for the recovery of the control-flow data.
	\item Chapter \ref{ch:Preprocessing} describes what steps are needed to feed the extracted data to the neural network.
	\item Chapter \ref{ch:Struct2vec} presents the parameter settings and broadly explains the behaviour of structure2vec.
	\item Chapter \ref{ch:Evaluation} evaluates the performance of the neural network and what changes had to be done in order to perform the classification.
	\item Chapter \ref{ch:Summary} summarizes this thesis project and outlines potential future work.
\end{itemize}

\chapter{Background} \label{ch:Literature}
In this chapter I detail the required background to put this thesis project into context and understand the methodology used. I present the malware features that can be used for machine learning algorithms in Section \ref{sec:machine_learning} and different malware detection methods in Section \ref{sec:malware_detection}. Afterwards, I describe the different machine learning types in Section \ref{sec:machine_learning} and later in Section \ref{sec:control_flow_data} I explain the different control-flow data types used as features in this thesis project.

\section{Malware detection}\label{sec:malware_detection}

The traditional malware identification method is finding a sequence of bytes that is unique enough to detect a given sample \cite{aslan2020survey}. These sequences of bytes are usually called signatures. The creation of these signatures can be time-consuming because the anti-malware vendors must find such sequence, that identifies the given malware but does not cause false positives for other benign programs.

Heuristic-based methods, e.g., YARA rules\footnote{\url{https://yara.readthedocs.io/en/stable/} last accessed: 09/12/2021}, contain rules or patterns that are general enough to identify variants of the same malware, but still do not apply to benign files. Identifying these patterns requires experts, who are able to determine the middle ground between excessive false positives and inadequate true positives.

In order to adapt to the increased malware development, anti-malware vendors started utilising the computing capacity of the cloud. In this approach, any new file is compared to a local signature database. If this approach does not yield a detection, the file is uploaded to the cloud for further processing. After the analysis in the cloud, the results are sent back to the client. The client acts in accordance with the results. Because cloud computing power increased extensively in recent years, new methods are available to detect malware, including data mining in combination with machine learning \cite{A_Survey_on_Malware_Detection_Using_Data_Mining_Techniques}.

\section{Machine learning}\label{sec:machine_learning}
In order to detect malware using machine learning, analysed samples must be represented using specialised attributes, so called features. In the case of malware detection, these features are used by machine learning algorithms for classification into two (e.g., malicious or benign) or more (e.g., malware families)  categories. For their extraction, there are two main methods: static analysis and dynamic analysis \cite{The_rise_of_machine_learning_for_detection_and_classification_of_malware,Survey_of_machine_learning_techniques_for_malware_analysis}. 

\label{sec:static_analysis}
Static analysis is extracting features without actually running the given sample. Its target can be binary or source codes. For the purpose of analysis, binary files first must be disassembled or decompressed. Such disassembler programs are Ghidra\footnote{\url{https://ghidra-sre.org/} last accessed: 03/12/2021} and IDA\footnote{\url{https://hex-rays.com/ida-pro/} last accessed: 03/12/2021}. Although, these have their own limitations. Some malware apply countermeasures against disassembling, like compressed parts of the code, which is only decompressed runtime \cite{A_Survey_on_Malware_Detection_Using_Data_Mining_Techniques}.
Static analysis can also miss out on some niche parts of a malware, like event-based activation. Typical features extracted using static analysis include the following:

\begin{itemize}
	\item API calls: These are special function calls sent to the operating system by the program. Their type or order can help in identifying malware and underlying suspicious behaviour.
	\item N-grams: N-grams are N length byte sequences in the program. For example, the sequence "34532" contains 3 3-grams: "345", "453" and "532".
	\item Strings: In the program's code, raw strings can be used to identify websites or IDs. These can help in recognizing the attacker's intention.
	\item Opcodes: In assembly, each instruction has an operational code and single or multiple operands. These operational codes are often referred to as opcodes and sequences of them can help in the identification of a program.
	\item Control-flow graphs: All possible execution of a program can be represented in a directed graph format. Here each node is a code sequence without any condition and branching edges from a given node are the different results of the condition.
\end{itemize}

\label{sec:dynamic_analysis}
In comparison with static analysis, dynamic does execute the code in some kind of controlled environment. With the help of these, we can monitor the behaviour of malware without risking the infection of the analysis device.

\begin{itemize}
	\item Debugger: With the help of debuggers, malware can be analysed at instruction level. However, almost all malware have become able to recognise the presence of a debugger.
	\item Simulator, emulators: These observe the actions of executed malware in a controlled environment. For example, in virtual machines, the entire computer is emulated, including the hardware layer. Due to its all-round computer emulation, it can be quite resource-intensive, but offer an easy method to backup and restore the testing environment. Dynamic features acquired from these analysis can be e.g., execution trace, created tasks and processes, and modified files.
\end{itemize}

These features are used as input for the ML algorithm, which has two steps: model building and model usage. When building the model, the program optimizes how it should weigh each feature value when making a decision. The goal can be either classification or clustering the samples.

The aim of classification is to try to categorize the inputs into separate groups. This belongs to supervised learning because we know the possible outcomes. When building the classifying model, the data must be labelled, as the optimization is optimizing the model based on the successful label prediction.

In case of clustering, the data is unlabelled, and the goal is to group the similar inputs together without knowing what the output will be. This allows for example malware family categorization and automatic signature generation for these groups.

\section{Control-flow data}\label{sec:control_flow_data}
There are two types of control-flow data that are relevant to my work, control-flow graphs and call graphs. Both can represent possible executions of a program, although control-flow graphs do in a much more detailed way. Its advantage is obviously the more accurate representation, but also the potential to transform it into a call graph. For illustration, I wrote and compiled a simple C++ program (see Code \ref{code:C++_code}), then extracted its call graph and control-flow graph.

\begin{figure}
	\begin{lstlisting}[language=C++,caption={Simple example code.},label=code:C++_code,basicstyle=\normalsize,columns=fullflexible]
		void foo() {
		}
		
		void bar() {
			foo();
		}
		
		int main() {
			foo();
			bar();
			return 0;
		}
	\end{lstlisting}
\end{figure}

\begin{figure}[htb]
	\centering
	\includegraphics[width=\figuresize\textwidth]{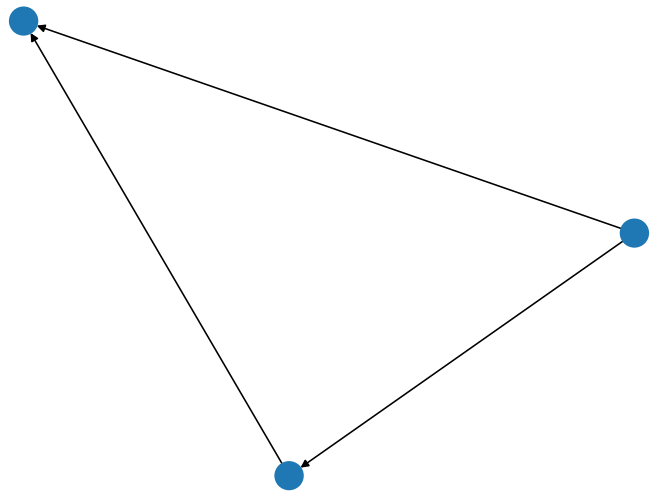}
	\caption{The generated call graph of Code \ref{code:C++_code}.}
	\label{fig:call_graph}
\end{figure}

In a call graph, the nodes represent functions and the directed edges the different calls between them. In Figure \ref{fig:call_graph}, the \normalcode{main()} function can be seen on the right, the \normalcode{foo()} at the top and the \normalcode{bar()} at the bottom.

\begin{figure}[htb]
	\centering
	\includegraphics[width=\figuresize\textwidth]{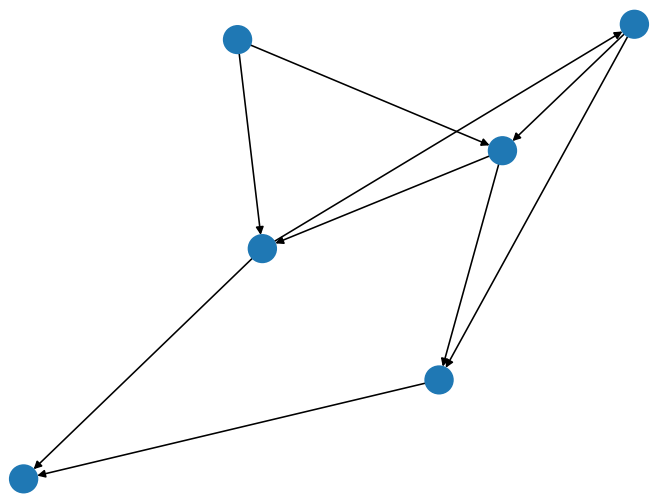}
	\caption{The generated control-flow graph.}
	\label{fig:cfg_fast}
\end{figure}

As mentioned before at Section \ref{sec:static_analysis}, control-flow graphs belong to static analysis. In this case, the nodes are sequence of compiled code lacking any transition. Two nodes are only connected with a directed edge if there is a jump instruction in the compiled code, be it conditional or unconditional. A control-flow graph contains not only the function calls, but also for example every if-else branch, or anything that contains a jump instruction in the compiled code. In Figure \ref{fig:cfg_fast} it is clear, that this resulted in a more complex graph as it is expected.

\section{Structure2vec}\label{sec:struct2vec}
When I was searching for previous works about machine learning algorithms for learning graphs, I found that most of them were designed for graph completion, such as finding missing edges and nodes. The only graph classification work I found was about an algorithm called structure2vec \cite{Struct2vec}. It is a neural network for classification based on labelled graphs. I used its PyTorch implementation\footnote{\url{https://github.com/Hanjun-Dai/pytorch\_structure2vec} last accessed: 09/12/2021}, which the authors made available on GitHub. The authors claimed that for large datasets it runs "2 times faster and produces 10 000 times smaller model than previous works while achieving state-of-the-art predictive performance" \cite{Struct2vec}. Because it was designed for molecule classification based on their graph structure, the similarity between this and my thesis project implied that it worth testing for control-flow graph classification.

\chapter{Overview}\label{ch:Overview}

In this chapter, I am going to discuss the malware detection framework's workflow at a high level, as well as the dataset I used during this thesis project. Section \ref{sec:workflow} discusses the framework's workflow. Section \ref{sec:dataset} presents the dataset and the data collection method.

\section{Workflow}\label{sec:workflow}
My goal was to create a proof of concept for machine learning based malware detection, using control-flow data as the input feature. The input of the workflow is a binary sample, which goes through a number of processing steps before it is classified as either malware or benign. I extracted control-flow data on a virtual machine and I implemented the rest of the framework on my home desktop PC. My PC has a Ryzen 5 2600 CPU, 16 GB of RAM, and a GTX 1060 GPU with 6 GB of VRAM.

\begin{figure}[!h]
	\centering
	\includegraphics{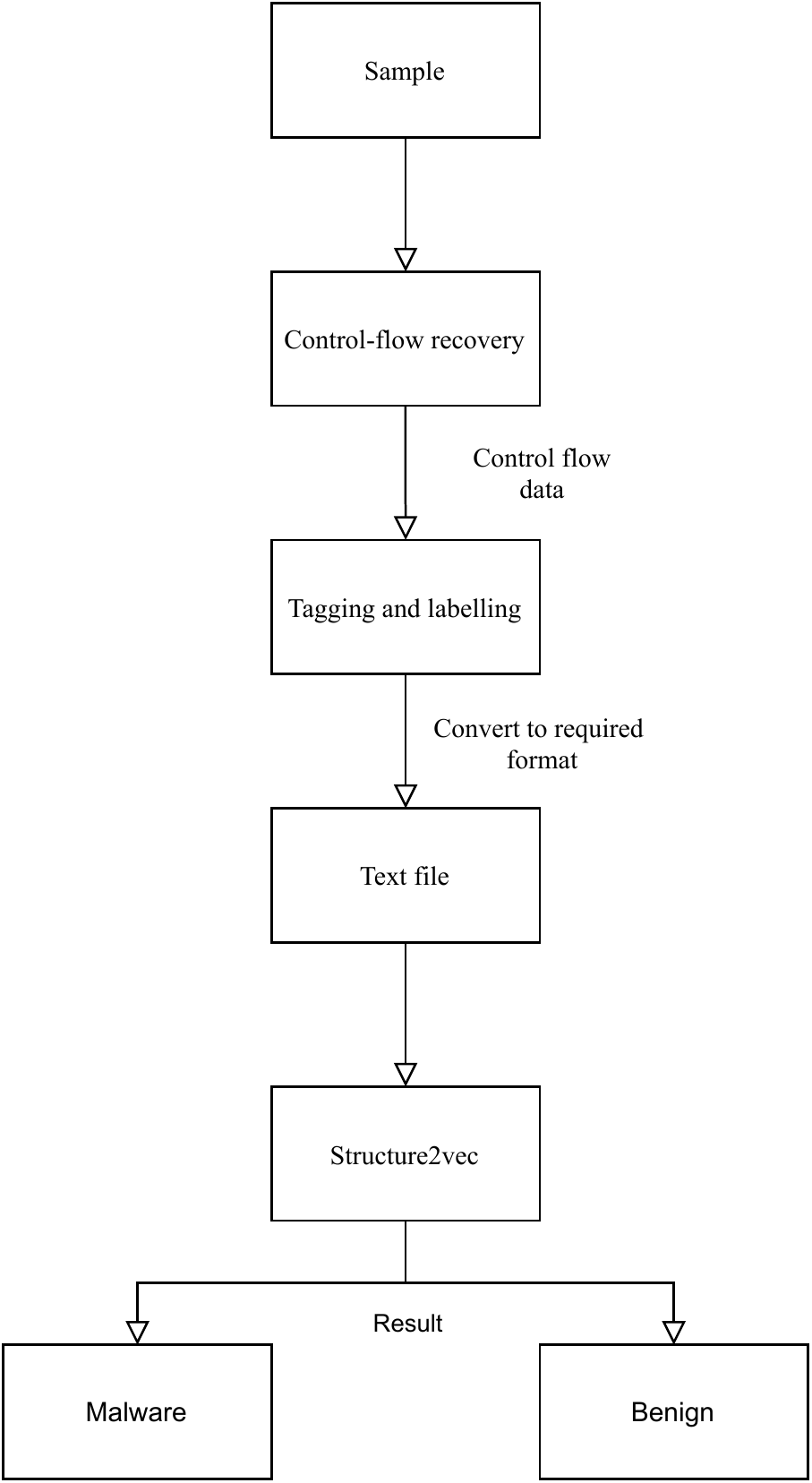}
	\caption{The workflow of my thesis project.}
	\label{fig:workflow}
\end{figure}

The overview of the workflow is shown in Figure \ref{fig:workflow}.  The first step is the control-flow recovery of the sample. Control-flow recovery results in a control-flow graph, which I use as the feature representing the sample. To implement control-flow recovery, I use the open source tool called angr\footnote{\url{https://angr.io/} last accessed: 24/11/2021}, which is a binary analysis framework. This tool can analyse both statically and dynamically linked binaries. I discuss control-flow recovery in detail in Chapter \ref{ch:CF_recovery}

In the second step, the graph tagging takes place. This tagging is required by the implementation of structure2vec, which uses this information to differentiate between the nodes of the input graphs. In order to generate tags for control-flow graph nodes, I assign unique numeric identifiers to the nodes based on the byte-sequences they represent. I present the implementation of tagging and structure2vec's required text input format in Chapter \ref{ch:Preprocessing}.



The last step of the workflow is the usage of the structure2vec algorithm in order to classify control-flow graphs as malware or benign. Although the original application of structure2vec is for chemical compounds and proteins, the algorithm itself is designed to classify graphs in general. The implementation of structure2vec can both train a neural network model and apply the trained model to the input graphs. Details of the used implementation is presented in Chapter \ref{ch:Struct2vec}.




\section{Dataset}\label{sec:dataset}
In order to train a neural network model for classifying control-flow graphs as either malware or benign a dataset of malicious and benign binaries is necessary. I received a set of malicious ARM binaries from my supervisor and I also had access to a set of benign ARM binaries. However, all of the benign binaries in the set are dynamically linked. This makes classification challenging because the neural network may notice that benign samples are always dynamically linked, while malicious binaries can be statically linked. Therefore, the neural network may base its decision on whether the control-flow graph contains library functions or not. 

To overcome this challenge, I needed the library dependencies of the benign binaries. These binaries were originally collected from the websites of two IoT vendors: D-Link and Ubiquiti. Therefore, I implemented a web crawler for both of their websites to collect firmware images and extract the required library dependencies.


\begin{figure}
	\includegraphics[width=\textwidth]{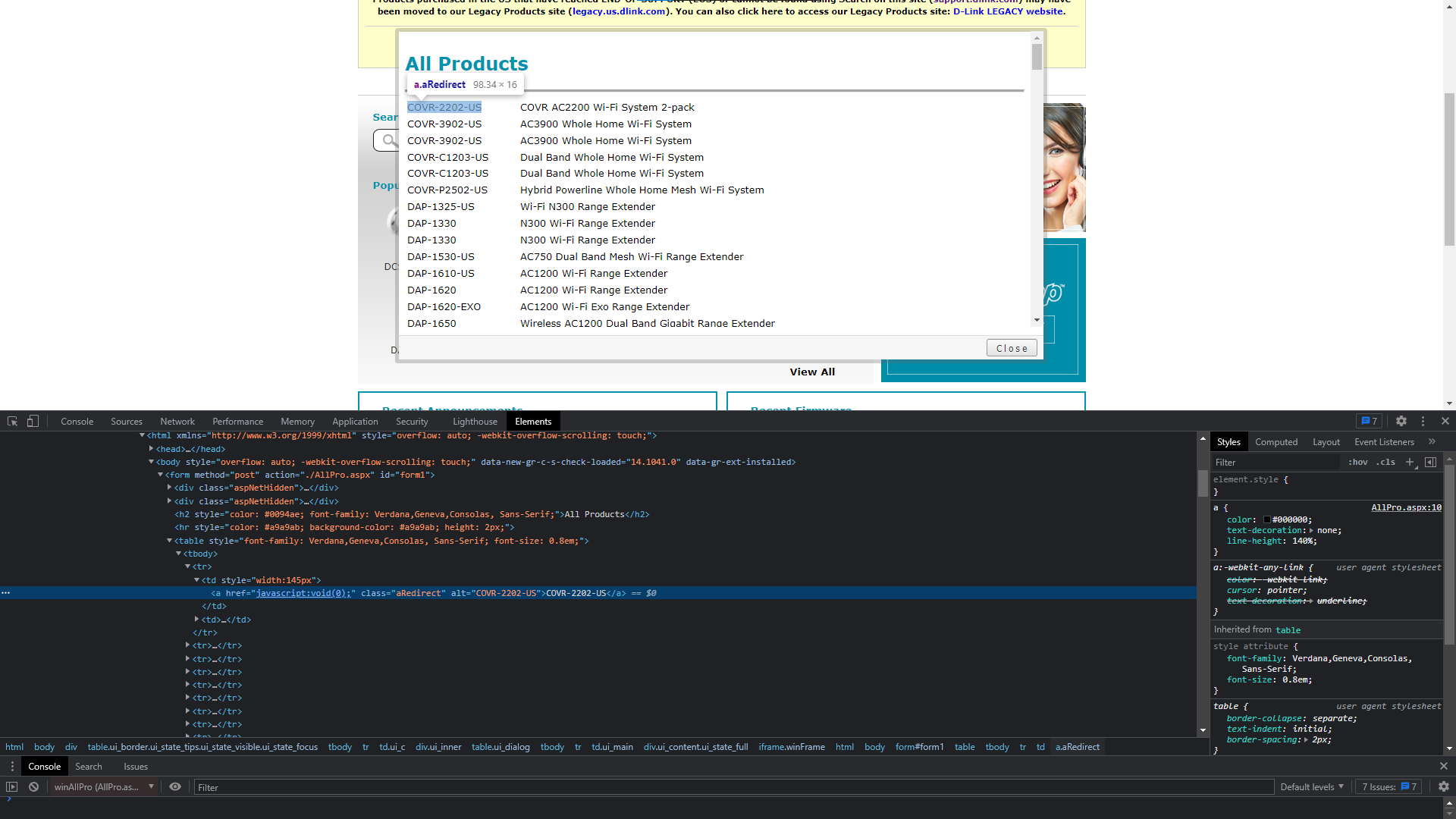}
	\caption{D-Link support page in developer mode.}
	\label{fig:dlink_page}
\end{figure}

First I created a web crawler for the D-Link support page\footnote{\url{https://support.dlink.com/} last accessed: 27/11/2021} in Python for automatizing the collection of links for firmware images. I used the selenium package\footnote{\url{https://pypi.org/project/selenium} last accessed: 09/12/2021} for searching and activating buttons on the website based on their xpath, attribute, class name and tag name. These can be seen in the browser's developer mode as seen in Figure \ref{fig:dlink_page}. After the collection of links, I filtered out links which pointed to irrelevant files like pdf spec sheets.

Most of the remaining files were in compressed format, therefore I first had to extract all of them. After, I needed to collect the executable binaries, and extract those with a Linux program called binwalk\footnote{\url{https://github.com/ReFirmLabs/binwalk} last accessed: 27/11/2021}. This tool recognises multiple file types and attempts to recursively extract every file from it. This resulted in a substantial amount of data and the required space for extracted firmware images increased from about 6-7 GB to more than 60. Next, I received an existing list of SHA-256 hashes of benign files to be used during model training and evaluation.

\begin{lstlisting}[language=Python,caption={Hash searching code snippet.},label=code:hash_search,basicstyle=\normalsize,columns=fullflexible]
	if file_hash in hash_list:
		print("Hash match found!")
		search_folder = f"{root.split(os.sep)[0]}{os.sep}{root.split(os.sep)[1]}"
		files_to_copy = list()
		for (c_root, c_dirs, c_files) in os.walk(search_folder):
			c_files: list[str]
			for c_file in c_files:
				if ".so" in c_file:
					files_to_copy.append(f"{c_root}/{c_file}")
\end{lstlisting}

For every found file, I needed its external libraries as all of them were dynamically linked. The script I created searched recursively in the root folder of the unpacked downloaded files. From there it copied each file if its name contained ".so" as these files are used as external libraries in Linux environment. Then the program automatically copied the required files to a different folder, categorized by their corresponding executable's hash. The only thing left was to add this folder's path for the angr analysis and generate the graphs for structure2vec to classify.

As machine learning algorithms work better with more data, I wanted to increase the benign file dataset, since I had a much larger amount of malware. Therefore, I also developed a crawler for additional benign samples from Ubiquity's download page\footnote{\url{https://www.ui.com/download/} last accessed: 27/11/2021}. After downloading the files, the process was exactly the same as with D-Link.

\chapter{Control-flow recovery}\label{ch:CF_recovery}

For the control-flow data recovery, I used an open-source framework called angr, which offers two types of analyses for the control-flow graph generation. One of them is the \code{\normalsize CFGFast}, which uses static analysis. According to its documentation, this is faster, needs far fewer resources but is not as thorough. The other analysis is the \code{\normalsize CFGEmulated}, which uses lightweight symbolic execution \cite{Taint} for the analysis of the sample. This causes longer analysis time and higher memory usage. I compare them in two aspects: their produced graphs attributes in Section \ref{sec:cfg_comparison} and their performance in Section \ref{sec:angr_performance}. In this chapter, I examine which of the two would be more appropriate to use for my thesis project based on these results.

\begin{lstlisting}[language=Python,caption={Example for creating angr project.},label=code:angr_proj,basicstyle=\normalsize,columns=fullflexible]
	proj = angr.Project('../dlink/' + sample,
	load_options={
		'auto_load_libs': True,
		'except_missing_libs': True,
		'ld_path': [f'{path_to_libraries}{sample}']
	},
	use_sim_procedures=False)
\end{lstlisting}

The first step to using angr is to create a project as seen in Code \ref{code:angr_proj}. When defining a project, the first parameter is the path to the executable binary. Then, in the \normalcode{load\_options} dictionary, there are several parameters, from which I used three. The first is the \normalcode{auto\_load\_libs} which is a bool variable that defines whether angr should automatically load external libraries. The second is the \normalcode{except\_except\_missing\_libs} which sets if the framework should throw an error when external libraries are missing. The third, the \normalcode{ld\_path} gives angr the path to the external libraries which can be used when extracting control-flow data. After \normalcode{load\_option} the \normalcode{use\_sim\_procedures} can be used to set if angr should simulate the behaviour of known functions during the analysis.

\begin{lstlisting}[language=Python,caption={Example for creating angr project.},label=code:analysis,basicstyle=\normalsize,columns=fullflexible]
	fast = proj.analyses.CFGFast()
	emulated = proj.analyses.CFGEmulated()
	
	fast_cfg = fast.graph
	emulated_cfg= emulated.graph
\end{lstlisting}

After the creation of angr project, two control-flow graph extraction analysis is available: \normalcode{CFGFast} and \normalcode{CFGEmulated}. As presented in Code \ref{code:analysis}, these techniques are available from the created angr project  \normalcode{analyses} attribute through functions with the same names. From the produced analysis, the created networkx graphs are available from their \normalcode{graph} attribute.

The goal was to improve malware detection for IoT devices, so the architecture of the sample database was ARM. It turned out that there are inputs from the malware database that causes crashes in the framework because of some angr bugs\label{ch:angr_bug} which I will discuss in Section \ref{sec:bugfixes}. First I had to filter which files could be analysed by both. \normalcode{CFGFast} was able to successfully analyze 393 files of the 1.061 available ARM malware samples. Of these 393 files, \normalcode{CFGEmulated} could only process 37 samples as the rest of the samples caused angr to crash.

\section{Comparison based on graph attributes \label{sec:cfg_comparison}}
Both \normalcode{CFGFast} and \normalcode{CFGEmulated} use the networkx\footnote{\url{https://networkx.org/} last accessed: 27/11/2021} package for graph creation, which is a widely-used Python package for graph-related workloads. From networkx, it uses the \code{\normalsize DiGraph} class, which means generated control-flow graphs are directed, they can contain loop edges and can not contain parallel edges. My first goal was to decide which technique to use. For this, I compared the two methods using the same input malware samples. Since all generated control-flow graphs were also networkx graphs, most of the graph attributes could be queried with networkx functions. These graph attributes were the following:

\begin{itemize}
	\item Number of nodes: \code{\normalsize networkx.number\_of\_nodes()}
	\item Number of edges: \code{\normalsize networkx.number\_of\_edges()}
	\item Number of components: \code{\normalsize networkx.number\_weakly\_connected\_components()}
\end{itemize}

For all of these, the inputs were the extracted control-flow graphs. Other attributes, such as the number of resolved jump kinds and the recognised system call nodes, were not available through networkx functions, as these are not general graph characteristics. Each control-flow graph node has an attribute called \code{\normalsize is\_syscall} which contains whether a node is a system call or not. In order to acquire a list of system call nodes, I iterated over the graph nodes and if it was a system call node, I appended it to a list. As for the resolved jump kinds, I also had to iterate over each node. However, here I used an angr function called \code{\normalsize get\_successors\_and\_jumpkind()}, which is a function of the \code{\normalsize CFGModel} class in the main angr project.

\begin{figure}[!h]
	\centering
	\includegraphics[width=\figuresize\textwidth]{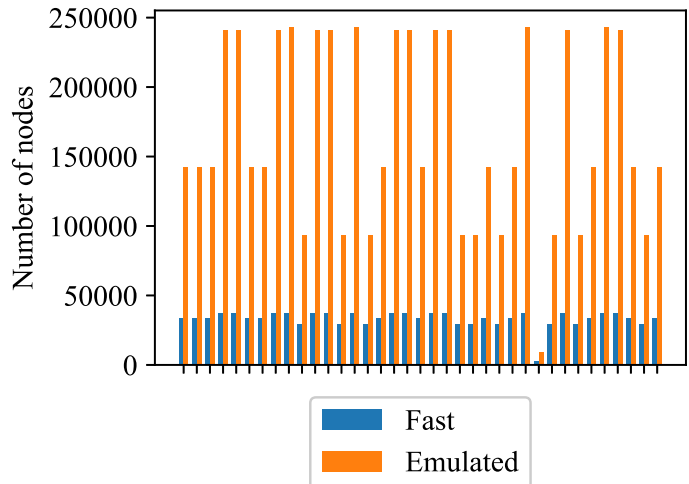}
	\caption{The number of nodes for each malware sample.}
	\label{fig:nodes}
\end{figure}

It is apparent that both methods gave a significant amount of information, though the emulated clearly resulted in a much larger amount of data, as can be seen in Figure \ref{fig:nodes}. The \code{\normalsize CFGEmulated} takes into account the call context which results in a more accurate representation, although its time complexity and memory requirement can increase exponentially. For this reason, I executed them with the simplest setup. However, it is possible that what it considered multiple nodes, could be in reality the same, only in different contexts.

\begin{figure}[htb]
	\centering
	\includegraphics[width=\figuresize\textwidth]{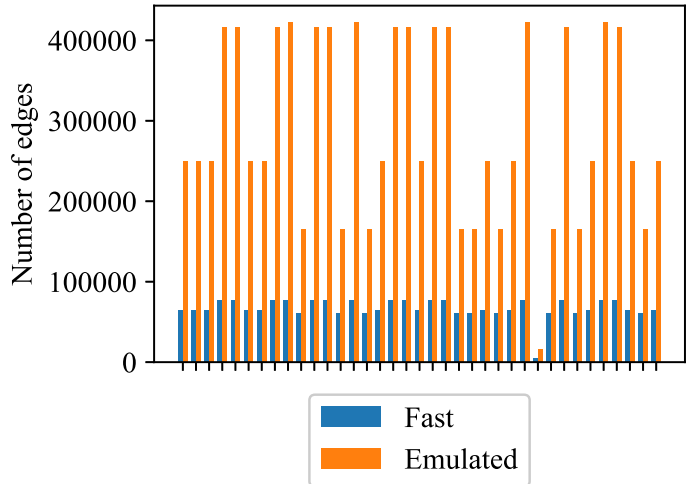}
	\caption{The number of edges for each malware sample.}
	\label{fig:edges}
\end{figure}

For the edges in Figure \ref{fig:edges}, the proportions are almost identical to the node analysis differing minimally from the node comparison in Figure \ref{fig:nodes}, with only the number being larger for all samples. As the in case of nodes, the context analysis may be the reason for the higher edge count for the \code{\normalsize CFGEmulated}.

\begin{figure}[htb]
	\centering
	\includegraphics[width=\figuresize\textwidth]{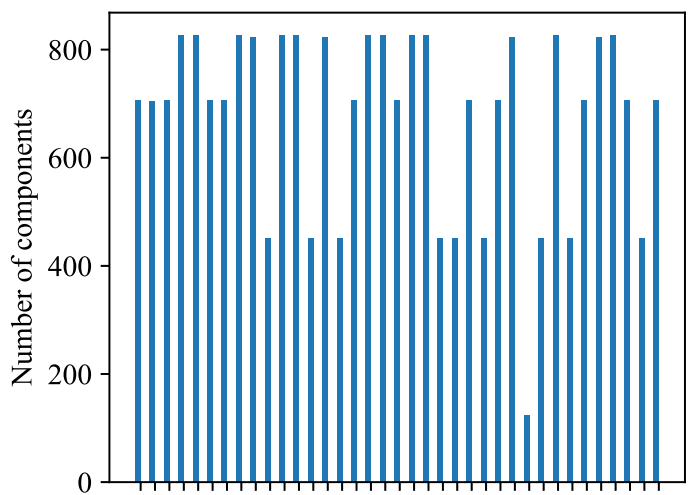}
	\caption{The number of components from \normalcode{CFGFast}.}
	\label{fig:components}
\end{figure}

In the case of the number of components, it is obvious that there are huge differences between the \code{\normalsize CFGFast} and \code{\normalsize CFGEmulated}. Although it is not visible, the \code{\normalsize CFGEmulated} found 1 component in all cases. Since the \code{\normalsize CFGEmulated} uses lightweight symbolic execution, it covers only those parts of the sample, which are are easily reachable from the entry point during analysis, e.g., do not require indirect jumps or the resolution of jump tables. This can explain the reason why it found exactly one component every time, as apparent in Figure \ref{fig:components}.

\begin{figure}[htb]
	\centering
	\includegraphics[width=\figuresize\textwidth]{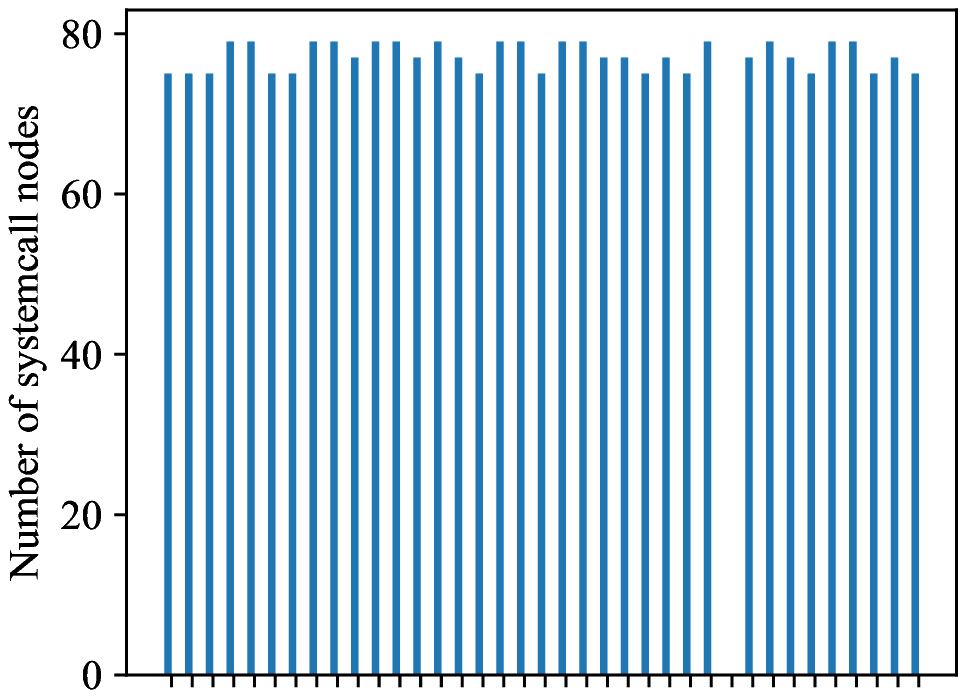}
	\caption{The number of system call nodes from \normalcode{CFGFast}.}
	\label{fig:syscall_nodes}
\end{figure}

The \code{\normalsize CFGEmulated} was not able to recognize the system calls only the entry and exit nodes, whereas the \code{\normalsize CFGFast} found tens of nodes containing system calls as it is visible in Figure \ref{fig:syscall_nodes}.

\begin{table}[htb]
	\centering
	\begin{tabular}{|lr|}
		\hline
		Only \normalcode{CFGFast}: & 2.011.846 \\
		\hline
		Only \normalcode{CFGEmulated}: & 2.824 \\
		\hline
		Both: & 2.454.102 \\
		\hline
		Neither: & 391.676 \\
		\hline
	\end{tabular}
	\caption{Number of covered instruction addresses.}
	\label{tab:addresses}
\end{table}

The last comparison is the memory coverage. In Table \ref{tab:addresses}, I present the combined sample addresses covered by only \normalcode{CFGFast}, only \normalcode{CFGEmulated}, both of them, and neither of them. I chose \normalcode{CFGFast} for future work as it provided much better coverage compared to \normalcode{CFGEmulated}. Although it is questionable how many of the covered instructions could be resolved, as in Figure \ref{fig:components} there are much more components in the \normalcode{CFGFast} analyses which can be caused by unresolved jumps. Still, as this gives much larger coverage it is a better representation of the sample.

\section{Comparison based on performance}\label{sec:angr_performance}
The \code{\normalsize CFGEmulated} was about ten times slower and required much more memory as the generated graphs were much larger. Their processing speed could be increased with the use of parallelization, but the lack of proper multithreading in Python makes its implementation complicated. Also, the parallel processing would mean, that the required memory would also increase, which is already a bottleneck when using \code{\normalsize CFGEmulated}. Furthermore, the lack of system call node identification in \code{\normalsize CFGEmulated} meant, that useful features were lost during the analysis. 

\section{Bugfixes}\label{sec:bugfixes}

As previously mentioned in Chapter \ref{ch:angr_bug} there were bugs in angr that significantly reduced the number of successful analyses. I focused on fixing two bugs, as these were enough to considerably increase the number of samples that could be analysed. One of them appeared while using \code{\normalsize CFGEmulated}, the other during \code{\normalsize CFGFast}. I detail these bugs in Sections \ref{sec:CFGE_bugfix} and \ref{sec:CFGF_bugfix} respectively.

\subsection{CFGEmulated fix} \label{sec:CFGE_bugfix}

First I fixed the \code{\normalsize CFGEmulated} bug as shown in Code \ref{code:CFGEm_bug}, which was found in \code{\normalsize cfg\_emulated.py}\footnote{\url{https://github.com/angr/angr/blob/c87311b01579670716ee4a5b40297e042f609f91/angr/analyses/cfg/cfg_emulated.py}} at line 1618:

\begin{figure}[htb]
	\begin{lstlisting}[language=Python,caption={The code without empty list checking caused errors during emulated analysis.},label=code:CFGEm_bug,basicstyle=\normalsize,columns=fullflexible]
		for block_id in pending_exits_to_remove:
			l.debug('Removing all pending exits to %#x since the target function %#x does not return',
				self._block_id_addr(block_id),
				next(iter(self._pending_jobs[block_id])).returning_source,
				)
	\end{lstlisting}
\end{figure}

When processing the binary, angr starts to create the control-flow graph nodes from the entry point based on the addresses reached. Meanwhile, it keeps a record of the other addresses on which the program can continue running at the end of processing each node. It interprets these addresses as blocks and puts these blocks into a list called \normalcode{self.\_pending\_jobs} which can be indexed with the \normalcode{block\_id}. After analysing a block, it checks whether this block, after connecting the graph, points to a block that has already been added to the graph. These are \normalcode{pending\_exits} that have already been processed. To do this, it iterates through the list to remove blocks that have already been processed. The problem that occurred during the analysis was that when processing the samples, there were no items left in the \normalcode{pending\_jobs} to process. So the \normalcode{next()} call on the iterator was invalid and the program crashed with an error message. Since this code snippet is just a debug message, it was sufficient to just create a separate else branch for the empty list.


\begin{figure}
	\begin{lstlisting}[language=Python, caption={My bugfix with empty list checking.},label=code:CFGEm_sol,basicstyle=\normalsize,columns=fullflexible]
		for block_id in pending_exits_to_remove:
			if self._pending_jobs[block_id] != []:
				l.debug('Removing all pending exits to %#x since the target function %#x does not return',
					self._block_id_addr(block_id),
					next(iter(self._pending_jobs[block_id])).returning_source)
			else:
				l.debug('Removing all pending exits to %#x since the target function does not return',
				self._block_id_addr(block_id))
	\end{lstlisting}
\end{figure}

\subsection{CFGFast fix}\label{sec:CFGF_bugfix}
Angr supports a variety of analyses and these often include different sets of conditions. To evaluate these, there are different contraint solvers for which claripy provides a communication interface. It basically deals with any values, even non-specific ones. During the analysis, a ValueSet type was given to the framework and was called to claripy, but the code to connect it to the ValueSet API was missing.

\begin{lstlisting}[language=Python, caption={Original code in \normalcode{balancer.py}.},label=code:CFGFast_bug,basicstyle=\normalsize,columns=fullflexible]
	@staticmethod
		def _min(a, signed=False):
		if not signed: bounds = backends.vsa.convert(a)._unsigned_bounds()
		else: bounds = backends.vsa.convert(a)._signed_bounds()
		return min(mn for mn,mx in bounds)
	
	@staticmethod
	def _max(a, signed=False):
		if not signed: bounds = backends.vsa.convert(a)._unsigned_bounds()
		else: bounds = backends.vsa.convert(a)._signed_bounds()
		return max(mx for mn,mx in bounds)
\end{lstlisting}

I had to add this so that the framework could get the correct attributes of the ValueSet. Here, in the \normalcode{balancer.py}, there were missing implementations of the function \normalcode{\_unsigned\_bound()} for \normalcode{ValueSet} types and these invalid calls caused the errors.
\begin{figure}
\begin{lstlisting}[language=Python, caption={Bounds value setting with signed and unsigned checking.},label=code:CFGFast_sol,basicstyle=\normalsize,columns=fullflexible]
	@staticmethod
		def _min(a, signed=False):
			if not signed:
				if isinstance(backends.vsa.convert(a), vsa.valueset.ValueSet):
					converted = backends.vsa.convert(a)
					bounds = [(converted.min, converted.max)]
				else:
					bounds = backends.vsa.convert(a)._unsigned_bounds()
			else:
				bounds = backends.vsa.convert(a)._signed_bounds()
			return min(mn for mn, mx in bounds)
	
	@staticmethod
		def _max(a, signed=False):
			if not signed:
				if isinstance(backends.vsa.convert(a), vsa.valueset.ValueSet):
					converted = backends.vsa.convert(a)
					bounds = [(converted.min, converted.max)]
				else:
					bounds = backends.vsa.convert(a)._unsigned_bounds()
			else:
				bounds = backends.vsa.convert(a)._signed_bounds()
			return max(mx for mn, mx in bounds)
\end{lstlisting}
\end{figure}
I extended the code with checks for instances of \normalcode{ValueSet} and setting its bounds with its properties based on the \normalcode{ValueSet} inner structure. Due to the lack of time and the possibly occurring errors, I assumed that the number of regions is always one.

After fixing these bugs, I successfully created both the \normalcode{CFGFast} and the \normalcode{CFGEmulated} control-flow graphs for 1.054 samples. Considering the huge speed impact, resource requirement difference, and the previous comparison results in section \ref{sec:cfg_comparison}, I chose the \code{\normalsize CFGFast} for my further work. 

The benign samples were all dynamically linked, so I had to include external libraries for the analysis. I set angr's load options according to the following settings:

\begin{itemize}
	\item automatically load the required libraries from the given path
	\item throw an exception if there are libraries missing
	\item the path where the libraries located
\end{itemize}

As the malware samples were all statically linked, they required no such settings and external libraries. The original idea was to use the control-flow graphs as the feature for classification, but these were much more complex and required an extreme amount of memory, which was not available. However, it is possible to extract call graphs from control-flow graphs. Fortunately, angr provides an easy method for this, as the analysis result has an attribute that contains exactly the call graph as a networkx graph. Although in this case, the node names are starting memory addresses in contrast to the control-flow graph's naming scheme, which uses low-level instructions. The call graph in this default generated format cannot be used for struct2vec, so it must be processed before serialization.

\chapter{Preprocessing}\label{ch:Preprocessing}

As previously mentioned, structure2vec requires that the graph nodes must have some sort of tagging. At first, I used the nodes starting memory addresses for testing the graph export. The export format was specified in the structure2vec's GitHub repository\footnote{\url{https://github.com/Hanjun-Dai/pytorch_structure2vec/tree/master/graph_classification/data} last accessed: 27-11-2021}. The file must be a text file, where:

\begin{itemize}
	\item the first line contains the number of graphs ($N$).
	\item The next $N$ number of blocks describe the structure of the graphs. Each block is built according to the following:
	\begin{itemize}
	\item Their first line contains the number of nodes ($n$) in the graph and the graph label separated by space 
	\item The following $n$ lines:
	\item every $i$th node (0 based) is described in the $i$th line of the block 
		\begin{itemize}
			\item These lines start with the tag of the current node ($t$) and the number of neighbours($m$)
			the following $m$ numbers in the $i$th row indicate the indices (0 based) of the $i$th node's neighbours
		\end{itemize}
	\end{itemize}
\end{itemize}

In order to achieve this structure, I have to create a tagging method that is practical for the neural network and generate node indices for each and every graph. First of all, the program creates globally an empty dictionary for these node features used for tagging. This contains the byte sequences and the tags assigned to them. The number of graphs is trivial, the program writes the length of the benign and malware lists combined. Due to the fact that structure2vec was designed for undirected graphs, the control-flow graph is changed with the networkx function \normalcode{to\_undirected()} called on itself and returns an undirected version of itself.

The next step is the node tagging, where I iterate over the control-flow graph nodes and generate a new 0 based numeric ID for every new byte sequence as seen in Code \ref{code:byte_sequence}. This is possible via the node's \normalcode{byte\_string} attribute from the angr framework.

\begin{figure}[htb]
	\begin{lstlisting}[language=Python, caption={Tagging of control-flow graph nodes.},label=code:byte_sequence,basicstyle=\normalsize]
		def cfg_node_tagging(_fast: angr.analyses.cfg.cfg_fast.CFGFast, _tag_dict):
		"""
		Extends the given dictionary with given angr fast project nodes
		
		:param _fast: angr fast project
		:param _tag_dict: dictionary of bytes to CFG node
		:return: extended dictionary of bytes to CFG node
		"""
		for _node in _fast.model.nodes():
		node_bytes = _node.byte_string
		if node_bytes not in _tag_dict:
		_tag_dict[node_bytes] = len(_tag_dict) + 1
		return _tag_dict
	\end{lstlisting}
\end{figure}

The next stage is writing the number of nodes, which can be acquired with a simple networkx function call: \normalcode{networkx.number\_of\_nodes()}. Then, based on which list contains the name of the sample, the program writes the label of the graph. This represents whether the input file is malware (0) or benign (1).

The only thing left to do is to create the node indices, which I store in two dictionaries to be able to get the node from the index and vice versa. This is only an iteration over the control-flow graph nodes while assigning the indices in the dictionaries.

The last part is iterating over the "index to node" dictionary writing the lines to the corresponding nodes. First, the current node's tag must be defined, which must be an integer. The control-flow graph node has an attribute that contains the byte sequence, which can be used for indexing the dictionary containing the node tags. After writing the node tag, with another networkx function (\normalcode{networkx.all\_neighbors()}) the program creates a list containing the neighbour nodes, then writes the length of this list. The last step is writing every neighbours' indices, which is done with the "node to index" dictionary.
\chapter{Structure2vec implementation}\label{ch:Struct2vec}

Although structure2vec was designed with chemical and biological graph classification in mind, as the base principle is graph classification, the task is the same. It can be used with different configurations, as there are multiple configurable input arguments:

\begin{itemize}\label{parameters}
	\item \textit{mode}: possible arguments: cpu/gpu, this decides if the neural network should run on GPU which is more appropriate for these calculations, provided it has enough memory
	\item \textit{gm} (default: \textbf{mean\_field}): possible arguments: mean\_field/loopy\_bp, this is the type of optimization used while training the neural network
	\item \textit{data} (default: \textbf{None}): the folder of used data
	\item \textit{batch\_size} (default: \textbf{50}): size of used minibatch
	\item \textit{seed} (default: \textbf{1}): input for python, NumPy and TorchPy randomizer
	\item \textit{feat\_dim} (default: \textbf{0}): dimension of node features
	\item \textit{num\_class} (default: \textbf{0}): number of classification classes
	\item \textit{fold} (default: \textbf{1}): number of folds
	\item \textit{num\_epochs} (default: \textbf{1000}): number of epochs
	\item \textit{latent\_dim} (default: \textbf{64}): dimension of hidden layers
	\item \textit{out\_dim} (default: \textbf{1024}): structure2vector output size
	\item \textit{hidden} (default: \textbf{100}): dimension of regression
	\item \textit{max\_lv} (default: \textbf{4}): max rounds of message passing
	\item \textit{learning\_rate} (default: \textbf{0.0001}): the rate of learning in neural network
\end{itemize}

The first step is loading the graphs. The program creates a list where the graphs will be stored, a label dictionary for the graph labels and a feature dictionary for node tags. As it iterates over the graphs, checks if the graph label is already in the dictionary. If it is not stored previously, the program maps the size of the dictionary as the ID for the label.

Next, it creates an empty networkx graph, which will be later extended with the elements of the loaded graph. As it loads the node's line that contains the node's tag, number of neighbours, and neighbouring nodes indices, it checks the node tag existence in the feature library and maps an integer the same way as with the graph label. However, the node tags are also stored in a separate list for every graph, which will be used later. Then two assertions happen, where it checks whether the number of imported edges and nodes are equal to those in the file. If the assertions were successful, the graph is added to a list, along with the list of node tags and the label of the graph wrapped in a separate class.

The next part is my addition to the code. I used scikit-learn\footnote{\url{https://scikit-learn.org/stable/} last accessed: 07/12/2021} for randomizing the dataset and splitting them into training and test lists.

After graph loading, the program sets whether the neural network should run on the GPU or the CPU and what kind of optimizer to use based on the starting arguments.

\chapter{Evaluation}\label{ch:Evaluation}

The evaluation revealed that the available memory would not be sufficient for the neural network, as it would have required roughly half a terabyte of memory even on the lowest parameters. Therefore, it was necessary to reduce the size of the input data in some way. I decided to use call graphs as a source for selecting the nodes from the control-flow graphs because the angr analysis already generates them. This change required a few modification in the previously mentioned tagging method. Angr provides a tool for searching for control-flow graph nodes, based on addresses. As the call graph nodes' names are their addresses, they can be used for searching the control-flow graph node, which contains the function call. The acquired control-flow graph node then can be used to assign the tag for the given call graph node.

The number of folds and epochs were irrelevant for my evaluation as I did not have a large amount of data. If I were to run multiple epochs, the neural network would overfit itself to the small database. This means that it would show 100\% accuracy, which is not representative when testing on new data.  Considering these, I decided to not use folding techniques and run multiple epochs, only 1 iteration.

\section{Runtime performance}

When I first executed the analysis, I left all the parameters at the default value, except mode and data, which I set to GPU and CALL respectively. This attempt was not successful, as it tried allocating approximately 68.5 GB, which was not only much more than my VRAM (6 GB) but even my system memory (16 GB). From this on, I set the execution type to CPU, while I tried to lower the memory requirements. By reducing the batch size I was able to successfully evaluate the structure2vec's performance. However, this still required more memory, than my GPU had and I had to execute on my CPU (AMD Ryzen R5 2600). Because of the lacking GPU memory, the speed performance was quite low. The training phase took 8:05 minutes and the testing phase 1:10, even with this low amount of data.

\section{Detection performance}

As I had much more malware samples than benign, I maximized the size of both datasets to the size of benign to achieve around a 1:1 ratio. This resulted in a training dataset with a size of 547 and a testing set of 137 which were both randomized before training and testing.

\begin{table}[htb]
	\centering
	\begin{tabular}{l|l|c|c|c}
		\multicolumn{2}{c}{}&\multicolumn{2}{c}{Predicted}&\\
		\cline{3-4}
		\multicolumn{2}{c|}{}&Positive&Negative&\multicolumn{1}{c}{Total}\\
		\cline{2-4}
		\multirow{2}{*}{Truth}& Positive & $68$ & $3$ & $71$\\
		\cline{2-4}
		& Negative & $3$ & $63$ & $66$\\
		\cline{2-4}
		\multicolumn{1}{c}{} & \multicolumn{1}{c}{Total} & \multicolumn{1}{c}{$71$} & \multicolumn{1}{c}{$66$} & \multicolumn{1}{c}{$137$}\\
	\end{tabular}
	\caption{Confusion matrix}
\end{table}

\begin{itemize}
	\item Precision: How much was truly malware from the samples considered malware. 
	$$Precision = \frac{T_p}{T_P+F_p}$$
	\item Recall/Detecion rate: How much was considered malware from the malware samples.
	$$Recall = \dfrac{T_p}{T_p+F_n}$$
	\item F1 score: The weighted average of precision and recall.
	$$F_1\ score = 2*\dfrac{P*R}{P+R}$$
	\item False alarm rate: Considered malware despite being benign.
	$$False\ alarm\ rate = \dfrac{F_p}{T_n+F_p}$$
	\item Accuracy: How much was classified correctly according to its actual type.
	$$Accuracy = \dfrac{T_p + T_n}{T_p+T_n+F_p+F_n}$$
\end{itemize}

From the measurements three of them are equal. Precision, recall, F1 score are 95.77\%, the false alarm rate is 4.55\% and the accuracy is 95.62\%.

\chapter{Summary and future work}\label{ch:Summary}

In this thesis, I presented the current state of embedded Internet of Things devices along with their security issues and the necessity of appropriate malware detection. I designed and implemented a proof of concept framework, which uses two types of control-flow data for malware detection. These extracted features are control-flow graphs and call graphs, which are used for classifying malware. The resulted framework achieved 95+\% detection and accuracy rate, while the false alarm rate was below 5\%.

Although the accuracy of classification is good, there is room for improvement regarding the speed of control-flow data recovery. Currently, about 6,5 hours are required for extracting the data from the used 1.396 portable executables. For considerably faster analysis, the optimization of angr is essential. The examination of control-flow data is a particularly complex task, hence other ways should be considered. Parallelization can considerably improve its performance hence it is a parallelizable problem, though its effectiveness can only linearly scale with the number of execution units.

There were also limitations considering the size of the dataset. For machine learning algorithms, a much larger dataset is required for thorough training and testing. I had to limit the malware and benign dataset to the benign set to able to train the neural network equally for malware and benign samples. It would be interesting to see how it would perform with a larger and more diverse dataset.

A possible work in the future is porting the neural network component to TensorFlow Lite\footnote{\url{https://www.tensorflow.org/lite} last accessed: 24/11/2021}. TensorFlow is one of the most popular and widely accepted machine learning frameworks besides PyTorch. Its beforementioned lite version is designed to work on IoT and other devices with smaller computational power. The implementation of this thesis's framework and performance evaluation in TensorFlow Lite deserves its own separate research.

In case the deployment to the embedded devices is not feasible, a client-cloud architecture is still achievable. This way the client processing power and capabilities are not present as limiting factor and the power of the cloud is applicable. For easier server installation the use of Docker\footnote{\url{https://www.docker.com/} last accessed: 27/11/2021} may be considered. There is minimal performance impact if any, and it reduces the installation of the entire framework to only execution of a few commands. Angr already has a dockerfile in its GitHub repository and the C++ implementation of structure2vec also has one. Considering these, it should be possible to create a merged dockerfile containing both parts connected to each other.

\chapter*{\koszonetnyilvanitas}\addcontentsline{toc}{chapter}{\koszonetnyilvanitas}

The presented work was carried out within the SETIT Project (2018-1.2.1-NKP-2018-00004), which has been implemented with the support provided from the National Research, Development and Innovation Fund of Hungary, financed under the 2018-1.2.1-NKP funding scheme.


\addcontentsline{toc}{chapter}{\bibname}
\bibliography{bib/mybib}


\end{document}